\documentclass{llncs}
\usepackage{graphicx}
\usepackage{amsmath,amssymb} 
\usepackage{float}
\usepackage{caption}
\usepackage{wrapfig}
\usepackage{subfiles}
\usepackage{subcaption}
\usepackage{color}
\DeclareCaptionFormat{citation}{%
   \ifx\captioncitation\relax\relax\else
     \captioncitation\par
   \fi
   #1#2#3\par}

\let\captioncitation\relax
\usepackage[width=122mm,left=12mm,paperwidth=146mm,height=193mm,top=12mm,paperheight=217mm]{geometry}
\begin{document}
\pagestyle{headings}
\mainmatter
\def\ECCV18SubNumber{***}  

\title{A Survey on Knowledge integration techniques with Artificial Neural Networks for seq-2-seq/time series models} 



\author{Pramod Vadiraja\inst{1} \and Muhammad Ali Chattha\inst{2}}
\institute{
\email{vadiraja@rhrk.uni-kl.de}
\and
\email{Muhammad\_Ali.Chattha@dfki.de}}

\maketitle

\begin{abstract}
In recent years, with the advent of massive computational power and the availability of huge amounts of data, Deep neural networks have enabled the exploration of uncharted areas in several domains. But at times, they under-perform due to insufficient data, poor data quality, data that might not be covering the domain broadly,  etc. Knowledge-based systems leverage expert knowledge for making decisions and suitably take actions. Such systems retain interpretability in the decision-making process. This paper focuses on exploring techniques to integrate expert knowledge to the Deep Neural Networks for sequence-to-sequence and time series models to improve their performance and interpretability.

\keywords{Deep Neural Network, TimeSeries, Sequence-to-Sequence models, Expert Knowledge}
\end{abstract}

\section{Introduction}
\textbf{Knowledge Based Systems}:\\
Systems that incorporate human expertise for making decisions are knowledge-based systems (KBS). Traditionally a knowledge-based system consists of a knowledge base which is data suitably collected and organised by human experts in various fields, inference engine - that relies on the knowledge base for decision making, a working memory to handle operations. The inference engine can be rule-based, case-based, etc.\\
\textbf{Deep Neural Networks}:\\
Deep neural networks, on the other hand, is more about statistical modelling that relies on massive amounts of data to find statistical patterns, non-linear relationships to be able to match the prediction patterns from a given training set. It relies on these patterns to infer conclusions about new data as well.\\ 
\textbf{Time-series models}:\\
A Recurrent neural network is a class of neural network that deals with the prediction of temporal sequences. Long-Short term memory (LSTM), Gated Recurrent Units(GRU) are some of the Recurrent neural network(RNN) architectures that are used for time series forecasting.\\\\
\textbf{Sequence to Sequence models}:\\
Sequence to sequence models aims to translate a fixed-length input sequence to a fixed-length output sequence where the length of the input and output may differ. It mainly has three parts: the encoder, intermediate vector and the decoder. In the encoder, several stacks of recurrent units (LSTM or GRU) are combined such that each unit accepts an input element from the sequence and propagates it, thus forming an intermediate hidden state. This information in the hidden state is consumed by the decoder part of the network that in turn consists of sequences of recurrent units that produce a sequence of outputs.
\subsection{Synergy between the Knowledge base systems and the Deep neural networks}
Although the Deep neural networks have shown promising performance in several fields, there exist areas like interpretability, reasoning in which they lack and hence needs attention. On the other hand, Expert Systems are built on top of the characteristics which the Deep neural networks lack. Hence there can be ways where we can leverage the strengths of both systems by various principles. This paper discusses some of the techniques of integrating expert knowledge to Deep Neural Networks to attain a kind of synergy between them.  

\section{Knowledge integration techniques}
\subsection{Time-Series models}
\subsubsection{}
Dou et al.\cite{PopularityPrediction} presents a technique for popularity prediction for online content over time, through the integration of expert content in the form of Knowledge Base (KB) information, with the time series data. The KB information (framed as a set of triples) is represented as an embedding vector that is learned by minimizing the loss of the triples, thereby capturing similarities in the triples. The technique also proposes to enhance the prediction for an item, through attention-based integration of KB neighbors (items with similar KB information). A LSTM network is used for predicting the popularity over time.\\
Using attention mechanism while considering similar KB neighbors provides the advantage of considering only similar behaviours of the neighbours for popularity predictions. Usage of KB embedding for the Knowledge graph enables the expert to intervene and update the Knowledge graph easily by adding triples to the graph which in turn can be captured while integrating with the time series data via the process of learning the KB embedding for the updated graph. 
\vspace{-3mm}
\subsubsection{}
Kilimci et al.\cite{DemandForecasting} presents a technique for Demand Forecasting for supply chains. They propose an integration strategy that is inspired by boosting the ensemble model in which decisions are made as a weighted combination of results. The two steps involved in the integration step are ensemble generation in which 11 forecasting models that includes forecasting algorithms along with Deep Neural Network and ensemble integration. Ensemble integration proposes a strategy(using weighted average) that not only focuses on the model that is performing consistently at its best currently but also in the previous weeks, as well as the previous year's transformation of the current and the previous week. This ensures that the forecasts are more reliable in terms of the trend and seasonality behaviours of the time-series signal.\\
However, because each model forecasts independently of each other, there might not be an actual exchange of expert knowledge from statistical time-series algorithms and the Deep neural networks.
\vspace{-3mm}
\subsubsection{}
Sadaei et al.\cite{LoadForecasting} presents a technique for Short-Term Load Forecasting(STLF), by the integration of the fuzzy time series (FTS) with convolution neural networks. This technique explores ways of translating multivariate time-series data into images of various channels and then passing the image to a Convolutional Neural Network (CNN) to capture meaningful features. The strategy uses FTS (using Chen uni-variate FTS algorithm \cite{Chen_Algo}) which is a sequence of fuzzy-sets along with the original time-series and converts them to the format of images. Each channel of the image corresponds to one of the variables of the overall time-series data. The strategy for the conversion is - For each variable of the signal, represent each value of the variable as a vector of 1s and 0s, with 1 in the vector corresponding to the position of the value in the sorted list of values. These vectors are horizontally stacked to form a matrix. Each variable of the signal will have a matrix, and each of these matrices corresponds to one of the channels in the final image.\\ 
This technique proposes a way to visualize the multi-variate time-series signals as an image that is encoded with components like the seasonality, trend. Because of the usage of the Fuzzy Time Series that in-turn uses fuzzy logic, any kind of expert knowledge can be encoded in the image. Using CNN strengthens the task of identifying the strong features in the input image to perform the prediction.
\vspace{-3mm}
\subsubsection{}
Deng et al.\cite{TCN} presents a technique for stock trend prediction and explanation. The structured knowledge in the form of a knowledge graph is integrated into the neural network as event embeddings. It also uses external knowledge from text corpus which are expressed as event tuples. These tuples are later linked with the knowledge graphs to increase the relevancy and introduce structure in the collected knowledge. The event embeddings are then combined with the time-series values and fed into the Temporal Convolutional Network (TCN). The TCN has a 1 dimensional filter that performs causal convolutions over the input to predict the next value at the current time step. The architecture of the proposed KDTCN is shown in Fig \ref{fig:m_figures}\subref{fig:TCN} cited from \cite{TCN}-Fig.2.\\
This technique does not consider the relevancy of the knowledge information with the input time series as much as the technique described in \cite{PopularityPrediction}. However, inclusion of text corpus data into the knowledge graph can be leveraged to expand the knowledge graph.
\vspace{-3mm}
\subsubsection{}
Dongxiao et al.\cite{WaveletTransform} presents forecasting technique based on wavelet transforms. A wavelet transform is a process of converting a signal in terms of a mother wavelet function that is subjected to scaling and translations. A Wavelet neural network is the network that takes the output of the wavelet transform operation and performs the task of forecasting. The coefficients (scaling and translating factors) corresponding to the decomposed constituents of the wavelet transform are used as a part of the hidden layer activations in the feed-forward neural network. This wavelet neural network is then combined with the expert knowledge to make improved predictions. The wavelet neural networks might not capture certain factors that are domain-specific, in which case the technique leverages on a set of rules established by experts. Based on the historical data, the logic formula from expert experience may be described as:
\begin{equation}
    forecastV(t) - firstV(t) + D \text{ x } firstV(t)
\end{equation}
where $firstV(t)$ is the first load value predicted by the neural network at time t; $D$ is the decision making factor at that moment; $forecastV(t)$ is the outcome obtained by the expert system predict at time t.\\
If the knowledge base in an expert system is enriched with better results, the prediction accuracy may be improved.
\subsection{Sequence-to-Sequence models}
\subsubsection{}
Bai et al.\cite{LearnSpellingFromTeachers} presents a sequence prediction technique for the task of automatic speech recognition(ASR). A pre-trained external language model(RNN) that acts as prior knowledge is integrated with the neural network. The prediction probability of the language model is fixed while the seq-to-seq network is being trained. The language model not only predicts the expected K hard labels but also provides an exhaustive list of soft labels which is used as prior knowledge by the seq-to-seq network to in turn improve its performance. The training step can be summarized by the following equations:
\begin{equation}
    L(\theta) = \lambda L_{CE}(\theta) + (1 - \lambda) L_{LST}(\theta)
\end{equation}
where $\theta$ is the parameters of the seq-to-seq network; $L(\theta)$ is the overall loss of the proposed technique; $L_{CE}(\theta)$ is the cross-entropy loss of the raw seq-to-seq encoder-decoder network; $L_{LST}(\theta)$ is the loss that is obtained by the Kullback-Leibler divergence(KLD) between the estimated probability of the language model(RNN) for the input sequence and the prediction probability of the seq-to-seq model; $\lambda \epsilon [0,1]$ is the weighted coefficient that controls the amount of contribution of each of the two losses considered in the final objective.\\
The non-involvement of the external language model during the testing phase decreases the complexity of the whole network. The overall loss combines the knowledge from the transcriptions and the knowledge from the language model. Hence they are able to incorporate external knowledge into the seq-to-seq network in comparison with similar fusion techniques like Shallow, Deep and Cold Fusion\cite{ColdFusion} without much of an overhead. As the probability distribution of the language model is an external aid to the encoder-decoder network, the method is flexible to accommodate any other language model.
\vspace{-3mm}
\subsubsection{}
Sriram et al.\cite{ColdFusion} presents a technique (Cold Fusion) that uses seq-to-seq model fused with external language model through a parametric gated function. The seq-to-seq model is trained along-side the pre-trained language model such that the hidden node of the language model's state is used by the seq-to-seq network, in turn, allowing greater flexibility in choosing which aspects of the language it needs to consider more at each step.
The training process can be summarized as :
\begin{equation}
h_{t}^{\mathrm{LM}} =\mathrm{DNN}\left(\ell_{t}^{\mathrm{LM}}\right) 
\end{equation}
\begin{equation}
g_{t} =\sigma\left(W\left[s_{t} ; h_{t}^{\mathrm{LM}}\right]+b\right)
\end{equation}
\begin{equation}
s_{t}^{\mathrm{CF}} =\left[s_{t} ; g_{t} \circ h_{t}^{\mathrm{LM}}\right]
\end{equation}
\begin{equation}
r_{t}^{\mathrm{CF}} =\operatorname{DNN}\left(s_{t}^{\mathrm{CF}}\right)
\end{equation}
\begin{equation}
\hat{P}\left(y_{t} | x, y_{<t}\right) =\operatorname{softmax}\left(r_{t}^{\mathrm{CF}}\right)
\end{equation}
where $h_{t}^{\mathrm{LM}}$ is the language model's probability; $\mathrm{DNN}$ can be a Deep neural network with any number of layers; $\ell_{t}^{\mathrm{LM}}$ is the language model's (LM) logit at time step $t$; $g_{t}$ the output of the gating function; $s_{t}$ state of the task specific model(seq-2-seq model); $s_{t}^{\mathrm{CF}}$ is the final fused state;
$\circ$ represents element-wise multiplication; $\hat{P}\left(y_{t} | x, y_{<t}\right)$ is the final predictive distribution of the fused network given the input sequences.\\
One of the offerings of this technique is that using the probability $(h_{t})$ instead of the language model's actual state, it eliminates the complexities that might arise if there is any modifications in the hidden nodes. Hence any language model can be used here, with minimal changes. By fusing the hidden states directly with the states of the seq-to-seq network, a relatively higher relevancy is achieved enabling the seq-to-seq network to focus on the required aspects of the expert language model. However, in comparison with \cite{LearnSpellingFromTeachers} this technique has increased complexity.
\vspace{-3mm}
\subsubsection{}
Cho et al.\cite{MemoryControlForS2S} presents a technique (Cell Control Fusion) that is similar to \cite{ColdFusion}, but differs in the aspect of not just fusing the gated outputs of hidden states of the external language model (RNN) but also for the cell states. Hence, a LSTM flavor of the RNN is used as a sequence-to-sequence model in this technique. It proposes two methods - a) Fusing the language model by controlling memory cell state that can be summarized by:
\begin{equation}
\begin{aligned}
{h}_{t}^{\mathrm{LM}}=& \tanh \left(\boldsymbol {W}_{1} l_{t}^{\mathrm{LM}}+{b}_{1}\right) \\
{g}_{t}=& \sigma\left(\boldsymbol {W}_{2}\left[{c}_{t} ; {h}_{t}^{\mathrm{LM}}\right]+{b}_{2}\right) \\
{c}_{t}^{\mathrm{CCF}}=& {c}_{t}+{g}_{t} \odot {h}_{t}^{\mathrm{LM}} \\
{s}_{t+1}, {c}_{t+1}=& \mathrm{LSTM}\left(\text { input, } {s}_{t}, {c}_{t}^{\mathrm{CCF}}\right) \\
\hat{p}\left(y_{t} | y<t, x\right) &=\operatorname{softmax}\left(\boldsymbol{W}_{3} {s}_{t}+{b}_{3}\right)
\end{aligned}
\end{equation}
The gated state ${g}_{t}$ that controls the contribution of the LM, is fused with the existing cell state ${c}_{t}$ of the task specific sequence-to-sequence model. Here the $\mathrm{LSTM}$ function takes the previous  cell state as the new updated $\boldsymbol{c}_{t}^{\mathrm{CCF}}$ state instead of ${c}_{t}$ that contains additional context obtained from the external LM. 
b) Fusing the language model by updating cell and the memory states of the LSTM decoder of the seq-to-seq model. Here, the technique proposes to incorporate the strategy in \cite{ColdFusion} as well as the method (a). It suggests multiple flavors of implementing this approach, wherein the LSTM function of the decoder is given the language model updated/fused cell and the hidden state as inputs that is propagated further.\\
Fusing the states of the language model to the seq-to-seq decoder in a granular level of the memory cell state and the hidden state gives this technique an edge over other related strategies. This enables the model to have access to greater linguistic context of the LM while training. 
\vspace{-3mm}
\subsubsection{}
Zhang et al.\cite{PosteriorRegularization} presents a framework that enables integration of arbitrary prior knowledge sources with the neural network for the task of machine translation (MT). This approach closely adapts \cite{PosteriorRegularization_SimilarApproach} where the expert knowledge represents a set of arbitrary posteriors that are constrained on various manually set conditions. In case of MT, the constraints may correspond to Bilingual Dictionary that gives the mapping between the source and the target word, limit for length ratios of the source and the target words, etc. The integration of these posteriors is brought about in the objective (maximizing likelihood) where the model tries to minimize the KLD between the predictive distribution and the desired posteriors. The proposed approach in \cite{PosteriorRegularization} differ in the way the posteriors are represented. \cite{PosteriorRegularization_SimilarApproach} represents the constrained posteriors using expectations:
\begin{equation}
    Q=\left\{q(\mathbf{y}): \mathbb{E}_{q}[\phi(\mathbf{x}, \mathbf{y})] \leq \mathbf{b}\right\}
\end{equation}
where $\phi(\mathbf{x}, \mathbf{y})$ are the constraint features and $b$ is the lower bound for the constraints. There are some issues with this approach that is addressed by \cite{PosteriorRegularization} where the desired constraints  are modelled as a log-linear model that offers the flexibility to represent arbitrary prior knowledge sources as features. The overall objective of this approach can be summarized as:-
\begin{equation}
\begin{aligned}
& \mathcal{J}(\boldsymbol{\theta}, \boldsymbol{\gamma}) 
=& \lambda_{1} \mathcal{L}(\boldsymbol{\theta})-
& \lambda_{2} \sum_{n=1}^{N} \mathrm{KL}\left(Q\left(\mathbf{y} | \mathbf{x}^{(n)} ; \boldsymbol{\gamma}\right) \| P\left(\mathbf{y} | \mathbf{x}^{(n)} ; \boldsymbol{\theta}\right)\right)
\end{aligned}
\end{equation}
where $\mathcal{J}(\boldsymbol{\theta}, \boldsymbol{\gamma})$ is the overall objective that we want to maximize. $\lambda_1$ and $\lambda_2$ are the hyper-parameters to control preferences between the likelihood $\mathcal{L}(\boldsymbol{\theta})$ and the desired distributions;
\begin{equation}
    Q(\mathbf{y} | \mathbf{x} ; \gamma)=\frac{\exp (\gamma \cdot \phi(\mathbf{x}, \mathbf{y}))}{\sum_{\mathbf{y}^{\prime}} \exp \left(\gamma \cdot \phi\left(\mathbf{x}, \mathbf{y}^{\prime}\right)\right)}
\end{equation} represents the log-linear discrete representations of the prior knowledge.\\
This method describes a novel way to model the prior knowledge such that it can accommodate any arbitrary knowledge sources. This may prove to be useful for the technique discussed in the next section.
\vspace{-3mm}
\subsubsection{}
\begin{figure}[!ht]
\begin{subfigure}[b]{0.5\textwidth}
\centering
\includegraphics[width= 6cm]{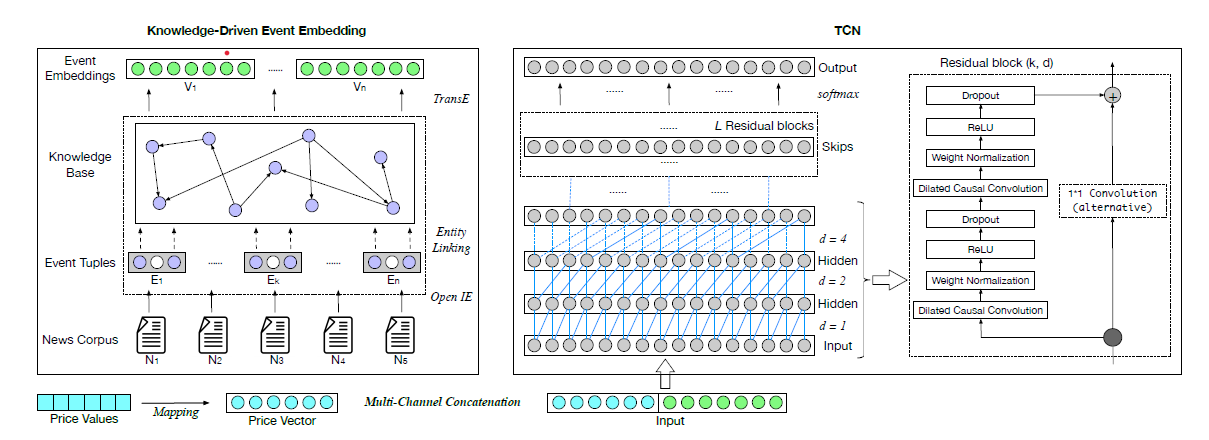}
\caption{}
\label{fig:TCN}
\end{subfigure}%
\begin{subfigure}[b]{0.5\textwidth}
\centering
\includegraphics[width= 6cm]{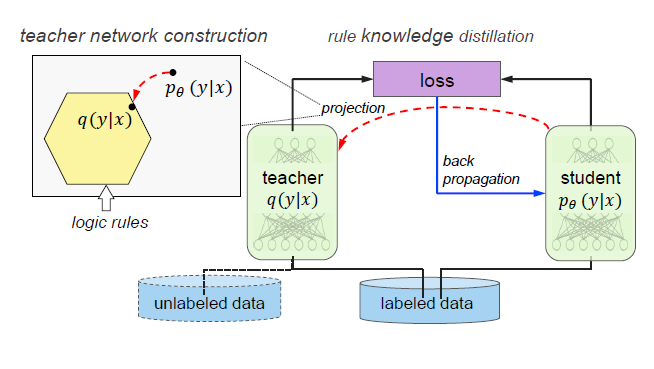}
\caption{}
\label{fig:TeacherNetwork}
\end{subfigure}%
\caption{}
\label{fig:m_figures}
\end{figure}

Hu et al.\cite{LogicRules} presents a technique - iterative distillation for transferring knowledge in the form of First Order Logic Rules to the weights of the neural networks. The architecture consists of a rule-regularized teacher network that trains the student network (CNN or RNN in general) to imitate the predictions of the teacher. The teacher network is obtained by projecting the student network to a rule-regularized subspace - this is done through adaptive posterior regularization ensuring that it is as close to the student network as possible and satisfies each of its rules, using KLD between the teacher and the student network. The architecture of the network is shown in Fig \ref{fig:m_figures}\subref{fig:TeacherNetwork} cited from \cite{LogicRules}-Fig.1. The red-arrow in Fig \ref{fig:m_figures}\subref{fig:TeacherNetwork} indicates projection of the student network into the rule regularised sub-space.
\vspace{-2mm}

Although the approach proposes a promising way of incorporating expert knowledge to the neural network, it is restrictive in terms of the amount of knowledge that can be incorporated. This is due to the limited a priori fixed constraints that are manually set by the experts. However, this can be taken care alongside the student training phase, by a learning procedure for the teacher networks and their weights jointly with the regulated deep neural network models\cite{MassiveLearnedKnowledge}.
\vspace{-3mm}
\subsubsection{}
\begin{figure}[!ht]
    \centering
    \includegraphics[height=4.4cm]{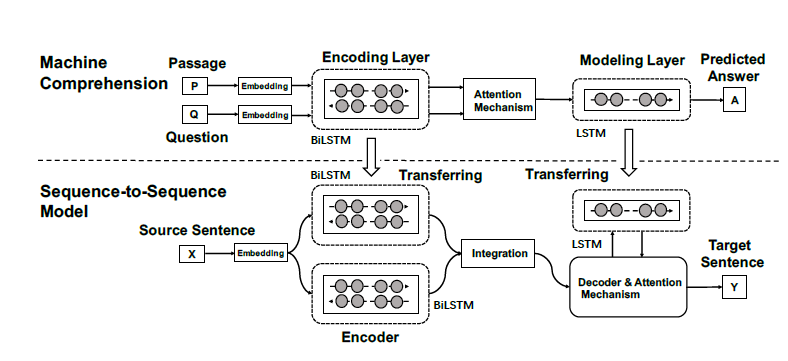}
    \caption{}
    \label{fig:MacNet}
\vspace{-3mm}
\end{figure}

Pan et al.\cite{MacNet} presents a technique that transfers knowledge from a Machine Comprehension(MC) architecture to a sequence-to-sequence model which in turn can be used in applications like machine translation, text summarizing. The MC architecture has an RNN based encoding layer, an attention layer and a modelling layer. The task-specific seq-to-seq network is an attention-based encoder-decoder network. With the ingestion of the MC model to this network enables the seq-to-seq model to capture deeper details and comprehend the content when performing the task of prediction. In the encoding stage of the task-specific model, the encoded result is concatenated with the encoded result of the MC model which in turn is used as the first input at the decoder stage of the seq-to-seq network. Assuming the seq-to-seq model is attention-based, at the decoding stage, its attention vectors are passed back to the MC's modelling layer that deeply captures the interaction of the source and the target states. The final prediction of the seq-to-seq network is computed as a weighted combination of the attention vectors from the MC model and the seq-to-seq network. The overall architecture of the proposed model is shown in Fig \ref{fig:MacNet} cited from \cite{MacNet}-Fig.1.
\vspace{-7mm}
Here, the MC model acts as a knowledge base that aids the sequence-to-sequence model. The huge learning curve of the MC model will give an edge to the seq-to-seq network while it predicts the output sequence. Any architecture can be used to implement the MC model. The framework can be applied to various NLP tasks. 
\vspace{-3mm}
\subsubsection{}
Long et al.\cite{DialogSystem} presents a conversational dialog system  that considers external knowledge related to the given query before composing a final response. Knowledge relevant to the current dialog history is extracted via a search engine using a set of keywords. A knowledge encoder is used to encode the extracted knowledge using a CNN followed by a max-pooling that is then delivered to the seq-to-seq model. The final goal is to generate informative responses to the input query incorporating external knowledge. This is done by augmenting the knowledge vector to the input of the decoder network as per the equations:
\begin{equation}
\begin{aligned}
x_{i}=\left[E\left(w_{r, i}\right), K_{v}\right] \\
h_{i}=L S T M\left(x_{i}, h_{i-1}\right) \\
p\left(w_{r, i+1}\right)=f\left(h_{i}\right)
\end{aligned}
\end{equation}
where $E$ denotes the word embedding, $K_{v}$ is the external knowledge encoded as a condensed vector; $f$ is the word projection layer that maps the hidden state to word distribution. It may be difficult for the model to compose noiseless responses by extracting relevant knowledge based on the input query.

\vspace{-3mm}
\section{Conclusions}
In this paper, we discussed some of the techniques for integrating expert knowledge in the form of First-order Logic Rules, tuples, embeddings etc with the neural network for time-series and sequence-to-sequence models. While each technique has its own set of pros and cons, it would be optimal to come up with a scalable technique that can incorporate the positive aspects of the above-discussed techniques.

\vspace{-3mm}
\bibliographystyle{splncs}
\bibliography{egbib}
\end{document}